\documentclass{bmvc2k}

\usepackage{booktabs}
\usepackage{amssymb}  
\usepackage{multirow}
\usepackage{varwidth}
\usepackage{floatrow}
\usepackage{wrapfig}

\newfloatcommand{capbtabbox}{table}[][\FBwidth]

\title{Back To The Drawing Board: Rethinking Scene-Level Sketch-Based Image Retrieval}

\addauthor{Emil Demić}{https://github.com/Emil-Demic}{1}
\addauthor{Luka \v{C}ehovin Zajc}{http://vicos.si/lukacu}{1}

\addinstitution{
 Faculty of Computer and Information Science\\
 University of Ljubljana\\
 Ljubljana, Slovenia
}

\runninghead{Demić, \v{C}ehovin Zajc}{Back To The Drawing Board}

\begin{document}

\maketitle

\begin{abstract}
The goal of Scene-level Sketch-Based Image Retrieval is to retrieve natural images matching the overall semantics and spatial layout of a free-hand sketch. Unlike prior work focused on architectural augmentations of retrieval models, we emphasize the inherent ambiguity and noise present in real-world sketches. This insight motivates a training objective that is explicitly designed to be robust to sketch variability. We show that with an appropriate combination of pre-training, encoder architecture, and loss formulation, it is possible to achieve state-of-the-art performance without the introduction of additional complexity. Extensive experiments on a challenging FS-COCO and widely-used SketchyCOCO datasets confirm the effectiveness of our approach and underline the critical role of training design in cross-modal retrieval tasks, as well as the need to improve the evaluation scenarios of scene-level SBIR. \\ Source code: \url{https://github.com/Emil-Demic/SketchScape}.
\end{abstract}


\section{Introduction}
\label{sec:intro}

In an exponentially growing multimedia content era, the ability to efficiently and intuitively retrieve relevant visual information has become an important challenge~\cite{datta2008image}. From personal photo organization to large-scale media archives and surveillance systems, there is a need for search methods that align with human intent (beyond rigid keyword matching)~\cite{lew2006content}. Recent breakthroughs in multimodal learning, particularly with vision-language models like CLIP~\cite{radford2021learning}, have demonstrated the power of leveraging large-scale language models to bridge the gap between semantic concepts and visual data. These models enable open-ended, natural language-based multimedia retrieval~\cite{jia2022scaling}. While textual queries offer flexibility, they may lack precision when users attempt to express complex spatial arrangements, fine-grained geometry, or visual attributes difficult to verbalize~\cite{chowdhury2022fscoco}. {\em Sketch-Based Image Retrieval} (SBIR) presents a compelling alternative; it offers a direct visual language for search, particularly valuable in creative and exploratory scenarios~\cite{eitz2012humansketch}.

However, most prior work in SBIR has focused on retrieving isolated objects or matching images at the category level~\cite{hu2011bag, yu2016sketch, yelamarthi2018zero, sain2021stylemeup, bhunia2020sketch, ling2022conditional}. {\em Scene-level} SBIR, which aims to retrieve entire photographic scenes based on free-hand sketches, poses a significantly more complex problem. This task demands models that encode information about composition, spatial relationships, and abstract scene semantics from a hand-drawn sketch, a sparse and often ambiguous query form, leading researchers to more complex task-specific models~\cite{liu2020sketcher, liu2022sketcherv2, chowdhury2022partially}.  

In our work, we return to the basics; we show that there is an untapped potential in the careful choice of pretraining, model architecture, and training objective formulation. Despite its simplicity, our proposed method achieves state-of-the-art results and significantly outperforms related work by doubling the retrieval performance on a challenging FS-COCO dataset~\cite{chowdhury2022fscoco}. We provide a detailed analysis and motivation of each design choice. Moreover, our analysis shows the impending limitation of current datasets that do not address the ambiguity of the data. We believe our insights will benefit further research in the field of SBIR, both in terms of method development and dataset creation.

\section{Related Work}
\label{sec:related}

Sketch-Based Image Retrieval is a challenging cross-modal task that requires matching sparse, abstract sketches to corresponding natural images. While early work in SBIR focused on category-level and instance-level retrieval~\cite{hu2011bag, cao2010mind, li2014deform, yu2016sketch}, recent research has increasingly shifted towards more complex and realistic scenarios, notably scene-level SBIR, where both query sketches and target images contain multiple semantically and spatially related objects~\cite {chowdhury2022fscoco, zuo2024scenediff, liu2022sketcherv2, chowdhury2022partially}.

Early approaches to scene-level SBIR, such as SceneSketcher~\cite{liu2020sketcher}, introduced graph-based models to encode spatial configurations among objects, demonstrating that scene layout information plays a crucial role in retrieval performance. Other approaches, such as SceneTrilogy~\cite{chowdhury2022fscoco}, explored incorporating text as an additional modality, enabling new capabilities while improving the generalization of purely sketch-based retrieval. More recently, generative models have been employed to bridge the sketch–photo modality gap more effectively. SceneDiff~\cite{zuo2024scenediff} introduced a diffusion-based framework that aligns sketches and images in a shared latent space, also incorporating text cues. This approach achieved state-of-the-art results, suggesting that generative alignment can offer more flexible representations for complex scenes. However, these models often rely on large models, dedicated architectures, and costly training procedures, raising concerns about scalability and reproducibility.

Recent advances in Sketch-Based Image Retrieval (SBIR) have also increasingly utilized multimodal architectures such as CLIP, owing to their strong generalization capabilities. CLIP-AT~\cite{sain2023clip} adapts CLIP through sketch-specific prompt learning, complemented by a regularization loss and patch-shuffling strategy to improve both category-level and fine-grained SBIR. MARL~\cite{lyou2024modality} takes a different approach by disentangling modality-agnostic semantics from modality-specific information. It achieves this by indirectly aligning sketches and images via contrastive learning with text, thereby constructing a more robust joint embedding space for cross-modal retrieval. More recently, SpLIP~\cite{singha2024elevating} introduces a bi-directional multimodal prompt-learning framework, further enhanced with an adaptive-margin triplet loss and a conditional cross-modal jigsaw task, achieving state-of-the-art performance in zero-shot SBIR.

A major bottleneck in advancing scene-level SBIR has, for a long time, lain in the lack of dedicated benchmarks. Most datasets used in early SBIR evaluation, such as Sketchy~\cite{sangkloy2016sketchy} and TU-Berlin~\cite{eitz2012humansketch}, were originally designed for category-level retrieval and contain primarily single-object sketches. SketchyScene~\cite{xu2018sketchyscene} and SketchyCOCO~\cite{gao2020sketchycoco} attempted to introduce scene complexity through compositional or template-based sketches, but they fall short in terms of sketch realism and diversity. FS-COCO~\cite{chowdhury2022fscoco} is the first large-scale dataset to offer freehand scene sketches paired with real-world MS-COCO~\cite{lin2014microsoft} images as well as accompanying captions. Despite its potential to support fine-grained, realistic evaluation of scene-level SBIR, it remains underutilized, with most methods exhibiting poor performance. In this work, we aim to close this gap by adopting FS-COCO~\cite{chowdhury2022fscoco} as a primary benchmark and demonstrating that a conceptually simple approach, with suitable architectural and training choices, can outperform more complex baselines.

\section{Methodology}
\label{sec:methodology}

Our methodology is driven by two key observations about scene-level SBIR. First, existing datasets are small and limited in diversity, making models prone to overfitting and emphasizing the need for strong pre-training and informed architectural choices that promote modality alignment. Second, sketch-image pairs are inherently ambiguous and noisy; sketches often omit or abstract key details. The learning objective must be robust to this uncertainty. To address these challenges, we build on pre-trained vision backbones and design a training objective that tolerates ambiguity while encouraging semantic consistency across domains.

\subsection{Model Design}

\begin{figure}
    \centering
    \includegraphics[width=\linewidth]{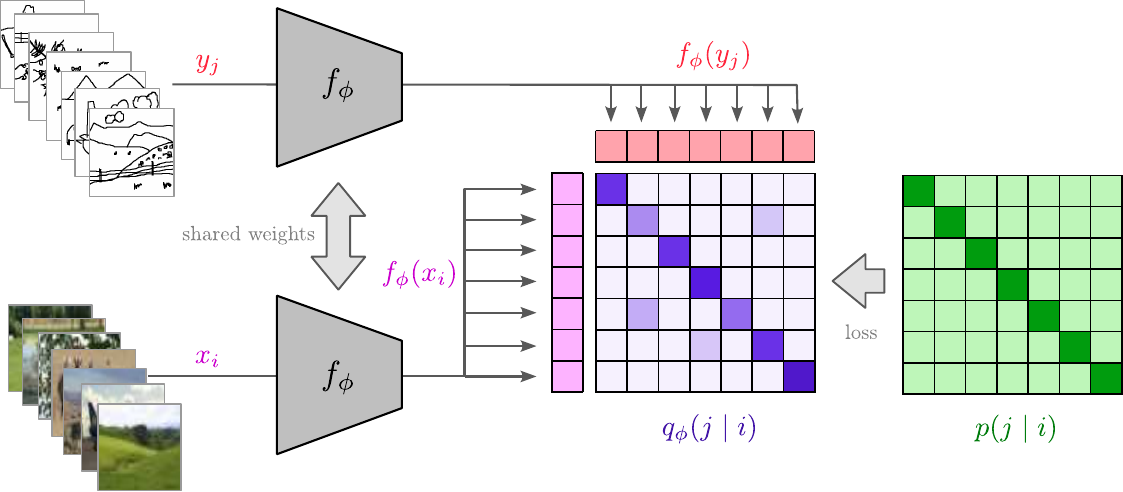}
    \caption{Overview of the proposed method. Our encoder model is trained in a Siamese manner, accepting both images and sketches. During training, the embeddings are matched on a batch level, aligning the similarity matrix with a target one. In the inference stage, the encoder is simply generating embeddings that are then compared using cosine distance.}
    \label{fig:method}
\end{figure}

As shown in Figure~\ref{fig:method}, we use an encoder network to project query sketches into an embedding space that aligns them with the corresponding images. We find that a shared-encoder Siamese architecture is very effective for scene-wide SBIR. Unlike natural language, which benefits from dedicated textual encoders, sketches are visual abstractions of the target modality. Using a single encoder enforces alignment in feature space and avoids overfitting to modality-specific characteristics, especially in situations where the amount of dedicated training samples is not sufficient. Despite the popularity of Vision Transformers~\cite{dosovitskiy2021image}, we choose ConvNeXt~\cite{liu2022convnet} due to its strong inductive biases for local structure and spatial hierarchies, which are particularly beneficial for capturing the compositional layout and fine-grained details common in sketches and natural scenes. Our detailed ablative experiments confirm this selection.

Due to small datasets, pre-training is essential in scene-level SBIR. Pre-trained models are typically trained on large image datasets, like ImageNet~\cite{deng2009imagenet}, for visual recognition tasks and are optimized to encode rich visual semantic concepts. In our work, we chose to use a model trained for cross-modal (text-image) alignment in the CLIP framework~\cite{radford2021learning} as a better fit for multi-modal alignment. Our experiments show that stronger visual-semantic priors enhance retrieval robustness in the presence of sketch ambiguity and improve generalization to complex, scene-level queries.

\subsection{Training Loss}

Most existing SBIR methods rely on triplet loss~\cite{schroff2015facenet} to learn sketch-image correspondences~\cite{li2023freestyleret, liu2022sketcherv2, chowdhury2022partially}, but this formulation has notable limitations due to its sparse supervisory signal and inefficient use of negative samples. Selecting informative negatives is particularly challenging in scene-level retrieval, where semantic overlap between scenes can obscure hard negative definitions. Inspired by advances in multi-modal representation learning, such as InfoNCE~\cite{oord2018representation} and CLIP~\cite{radford2021learning}, we adopt a contrastive learning setup where each sketch is paired with one positive and multiple implicit negatives from the batch. However, unlike InfoNCE~\cite{oord2018representation}, we account for the inherent ambiguity and noise in sketch queries, which increases the likelihood of accidental semantic alignment with supposed negatives. We treat the similarity scores across the batch as a probability distribution, $p(j \mid i)$, and align them with the supervision signal, $q_\phi(j \mid i)$, by minimizing the Kullback-Leibler (KL) divergence: 

\begin{equation}
\mathcal{L} = \sum_{i} \mathrm{KL}\left(p(\cdot \mid i) \,\|\, q_\phi(\cdot \mid i)\right) = \sum_{i} \sum_{j} p(j \mid i) \log \frac{p(j \mid i)}{q_\phi(j \mid i)},
\end{equation}

\noindent where indices $(i, j)$ denote data pairs. Similarly to InfoNCE loss~\cite{oord2018representation}, $q_\phi(j \mid i)$ is a \textit{learned distribution}, defined via a softmax over similarities between learned embeddings:
\begin{equation}
q_\phi(j \mid i) = \frac{\exp\left( \text{sim}(f_\phi(x_i), f_\phi(y_j)) / \tau \right)}{\sum_{k} \exp\left( \text{sim}(f_\phi(x_i), f_\phi(y_k)) / \tau \right)},
\end{equation}
where $\tau$ is a temperature hyperparameter and $\text{sim}(\cdot, \cdot)$ denotes the cosine similarity between embedding vectors of sketch $x_i$ and photos $y_j$ generated by the same model $f_\phi$. The \textit{supervisory distribution} $p(j \mid i)$ is defined as
\begin{equation}
p(j \mid i) = (1 - \alpha)\,\delta_{j=i^+} + \frac{\alpha}{N},
\end{equation}
\noindent where $i^+$ denotes the positive index, $N$ is the number of candidate samples, and $\alpha \in [0, 1]$ controls the degree of debiasing. This loss formulation introduces a debiasing mechanism that down-weights misleading negatives, and can be interpreted within the general Information Contrastive (ICon) framework~\cite{alshammari2025unifying}.

\section{Experimental Evaluation}
\label{sec:experiments}

We evaluate our approach on the FS-COCO~\cite{chowdhury2022fscoco} and SketchyCOCO~\cite{gao2020sketchycoco} datasets, and further conduct several in-depth studies to assess its performance. 
The model is implemented in PyTorch~\cite{pytorch} and built upon the CLIP~\cite{radford2021learning} architecture provided by the OpenCLIP~\cite{cherti2023reproducible} project. Specifically, we adopt the \textit{convnext\_base} variant pretrained on the \textit{laion400m\_s13b\_b51k} dataset. Only the visual encoder is retained, with all layers kept unfrozen. Optimization is performed using Adam~\cite{kingma2014adam} with a learning rate of $1\mathrm{e}{-4}$ and weight decay of $1\mathrm{e}{-5}$. The loss-related hyperparameters are set to $\tau=0.07$ and $\alpha=0.2$. Training was carried out on a single NVIDIA A100 40GB GPU with a batch size of 60 sketch–image pairs over 10 epochs, requiring approximately one hour. To improve efficiency, we utilize Automatic Mixed Precision (AMP) during training.

We follow the established protocol for fine-grained SBIR~\cite{yu2016sketch} and report \textit{recall at K} (R@K) for $K={1,5,10}$. This metric reflects the proportion of test queries for which the ground-truth image is ranked among the top-K retrieved results. 

\subsection{FS-COCO}

FS-COCO~\cite{chowdhury2022fscoco} is currently the only dataset offering human-drawn scene-level sketches. It contains $10,000$ scene-level images selected from the MS-COCO dataset~\cite{lin2014microsoft}, each accompanied by a corresponding freehand sketch as well as a text description. The sketches were drawn by $100$ non-expert individuals following a standard protocol. 

The dataset provides two evaluation splits, each consisting of $7,000$ training and $3,000$ testing samples. The first, referred to by the authors as the \textit{normal} split, allocates $70$ sketches per participant for training and the remaining $30$ for testing. The second, termed the \textit{unseen} split, assigns all $100$ sketches of $70$ participants to the training set, while using the full set of sketches from the remaining $30$ participants for testing. 

\begin{table}[h!]
\centering
\begin{tabular}{l|ccc|ccc}
             & \multicolumn{3}{l}{Normal} & \multicolumn{3}{l}{Unseen} \\
             & R@1     & R@5     & R@10   & R@1     & R@5     & R@10   \\
\hline
Siam-VGG~\cite{chowdhury2022fscoco}     & 23.3    & -       & 52.6   & 10.6    & -       & 32.5   \\
HOLEF-VGG~\cite{chowdhury2022fscoco}  & 22.8    & -       & 53.1   & \textit{10.9}    & -       & \textit{33.1}   \\
SceneTrilogy~\cite{chowdhury2023scenetrilogy} & 24.1    & -       & 53.9   & -       & -       & -      \\
SceneDiff (w Sketch)~\cite{zuo2024scenediff}    & 25.2   & 45.9   & 55.9  & -       & -       & -      \\
FreestyleRet~\cite{li2023freestyleret} & \textit{29.6}    & -       & \textit{56.1}   & -       & -       & -      \\
\textbf{Ours}         & \textbf{61.9}   & \textbf{81.4}   & \textbf{87.2}  & \textbf{60.0}   & \textbf{80.2}   & \textbf{86.1}     
\end{tabular}
\caption{Comparison of results on the FS-COCO~\cite{chowdhury2022fscoco} dataset for the normal and unseen splits. Best results in columns are presented in bold, second best in italic.}
\label{tab:fscoco_results}
\end{table}

We have compared our method to related work~\cite{chowdhury2022fscoco, chowdhury2023scenetrilogy, zuo2024scenediff, li2023freestyleret}, the results are summarized in Table~\ref{tab:fscoco_results}. As most of the reference methods do not provide their models, we report the results reported by the authors. Only the \textit{normal} split is used as a benchmark in most of the literature. Still, we can see that our approach outperforms all of the related work by more than double R@1 and follows a similar trend for more relaxed recall scenarios.

\begin{figure}[h]
    \centering
    \includegraphics[width=\linewidth]{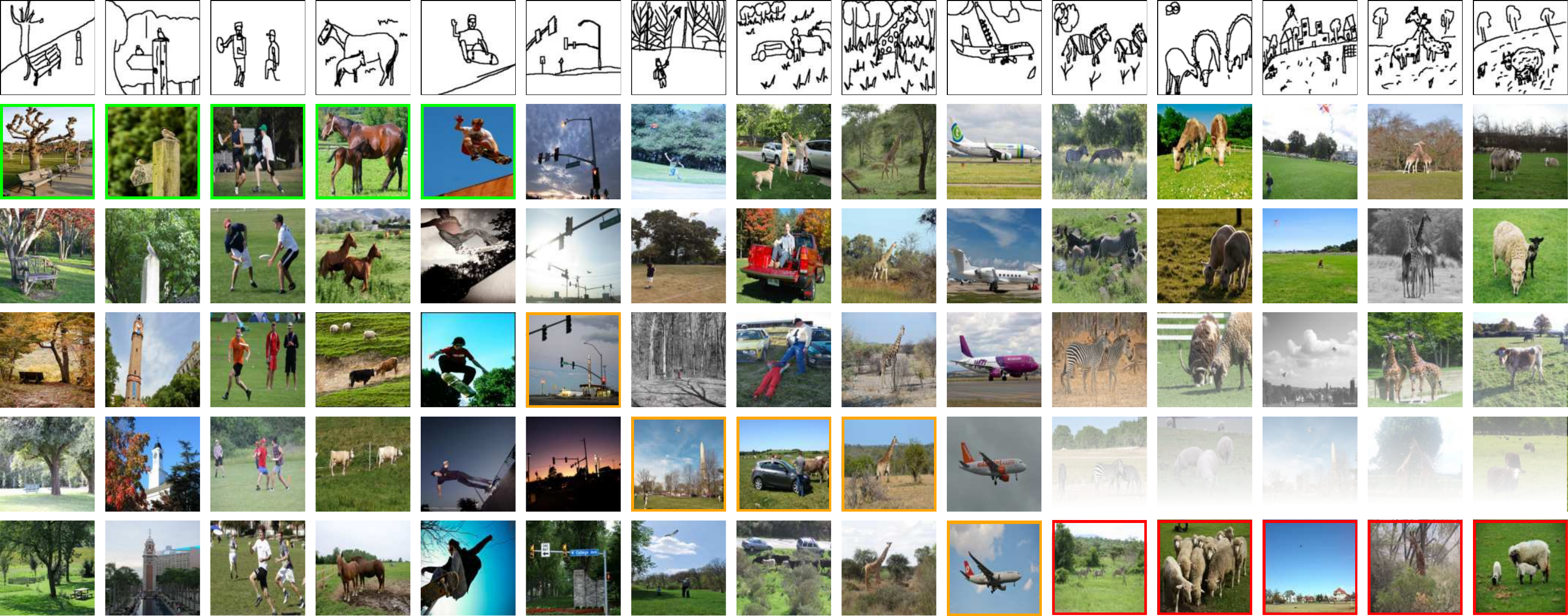}
    \caption{Qualitative results on the test split of the FS-COCO~\cite{chowdhury2022fscoco} dataset. Query sketches are shown in the top row, followed by the top matches. For the first five examples, the model correctly guessed the image; for the second five, the correct image was among the top five matches. The last five examples indicate failure cases where the correct match was not among the top ten matches; for reference, we attach the correct image at the bottom.}
    \label{fig:fscoco_results}
\end{figure}

Furthermore, we have evaluated our model on both proposed splits. We argue that the \textit{unseen} split better reflects real-world deployment, where models are unlikely to have access to numerous sketches from new users. The results in Table~\ref{tab:fscoco_results} show that this split is indeed more challenging and deserves more attention. Figure~\ref{fig:fscoco_results} contains several qualitative examples of the retrieval. We can clearly see that the dataset contains some very challenging subsets of very similar examples. In most failure cases (the right five examples), the method listed semantically very similar samples, which, considering the ambiguous nature of sketches, may not necessarily be considered a failure. This led us to investigate the embedding space of the model more thoroughly in Section~\ref{sec:embedding}.

\subsection{SketchyCOCO}

\begin{wraptable}{r}{0.6\linewidth}
\centering
\begin{tabular}{l|ccc}
                  & R@1   & R@5   & R@10  \\
\hline
Siam-VGG~\cite{chowdhury2022fscoco}          & 37.6  & -     & 80.6  \\
SceneDiff (w Sketch)~\cite{zuo2024scenediff}         & 34.3 & 69.1 & 81.4 \\
HOLEF-VGG~\cite{chowdhury2022fscoco}       & 38.3  & -     & 82.6  \\
SceneTrilogy~\cite{chowdhury2023scenetrilogy}      & 38.2  & -     & 87.6  \\
Partially Does It~\cite{chowdhury2022partially} & 34.5  & -     & 89.3  \\
SceneSketcherV2~\cite{liu2022sketcherv2}   & \textit{68.1} & \textit{87.6} & \textit{95.2} \\
\textbf{Ours}              & \textbf{70.0} & \textbf{90.0} & \textbf{96.2}
\end{tabular}
\caption{Comparison of results on the SketchyCOCO~\cite{gao2020sketchycoco} dataset. Best results in columns are presented in bold, second best in italic.}
\label{tab:sketchycoco_results}
\end{wraptable}

SketchyCOCO~\cite{gao2020sketchycoco} dataset comprises 14,081 sketch–image pairs, with images sourced from the MS-COCO~\cite{lin2014microsoft} dataset. It should be noted that due to the synthetic origin of sketches, they exhibit lower visual quality and realism compared to those in FS-COCO~\cite{chowdhury2022fscoco}. For this reason, most scene-level SBIR methods are evaluated on a subset, proposed in~\cite{liu2020sketcher}. In it, only sketches depicting more than a single foreground object are used. This yields a refined subset of $1,225$ sketch–photo pairs, $1,015$ for training and $210$ for testing. We adopt this evaluation protocol ourselves for comparison.

As presented in Table~\ref{tab:sketchycoco_results}, our method again achieves the best result at all three retrieval thresholds. This is especially impressive considering the fact that we do not use additional external data during training or inference. For comparison, other well-performing methods~\cite{liu2022sketcherv2,chowdhury2022partially} require supplementary instance segmentation annotations. This data adds additional context, which is usually not available and is not present in datasets like FS-COCO~\cite{chowdhury2022fscoco}.


\subsection{Ablation Analysis}
\label{sec:ablation}


To demonstrate the individual contributions of our method design choices, we have conducted an ablation analysis on the \textit{unseen} split of the FS-COCO~\cite{chowdhury2022fscoco} dataset. The results are summarized in Table~\ref{tab:ablation_results}.

\begin{table}[h!]
\centering
\begin{tabular}{lll|ccc}
Architecture                    & Loss                     & Pretrain & R@1   & R@5   & R@10  \\
\hline
\multirow{11}{*}{ViT-B}   & \multirow{2}{*}{/}   & ImageNet & 0.0   & 0.13  & 0.3   \\
                                &                          & CLIP     & 4.2   & 9.43  & 13.63 \\
                                & \multirow{3}{*}{triplet} & random   & 0.17  & 0.63  & 1.33  \\
                                &                          & ImageNet & 20.33 & 40.47 & 50.37 \\
                                &                          & CLIP     & 23.43 & 43.73 & 53.73 \\
                                & \multirow{3}{*}{InfoNCE} & random   & 0.17  & 0.90  & 1.87  \\
                                &                          & ImageNet & 32.33 & 54.87 & 65.17 \\
                                &                          & CLIP     & 50.93 & 73.20 & 81.73 \\
                                & \multirow{3}{*}{ICon}    & random   & 0.13  & 0.83  & 1.73  \\
                                &                          & ImageNet & 32.54 & 54.43 & 64.13 \\
                                &                          & CLIP     & 53.87 & 76.57 & 83.53 \\
                                \hline
\multirow{11}{*}{ConvNeXt-B} & \multirow{2}{*}{/}       & ImageNet & 0.03  & 0.23  & 0.47  \\
                                &                          & CLIP     & 6.37  & 14.7  & 21.23 \\
                                & \multirow{3}{*}{triplet} & random   & 0.07  & 0.33  & 0.73  \\
                                &                          & ImageNet & 36.23 & 57.10 & 66.33 \\
                                &                          & CLIP     & 22.73 & 42.90 & 54.23 \\
                                & \multirow{3}{*}{InfoNCE} & random   & 0.03  & 0.33  & 0.47  \\
                                &                          & ImageNet & 53.03 & 74.53 & 81.47 \\
                                &                          & CLIP     & 52.20 & 74.30 & 81.73 \\
                                & \multirow{3}{*}{ICon}    & random   & 0.07  & 0.13  & 0.30  \\
                                &                          & ImageNet & 54.53 & 74.63 & 82.00 \\
                                &                          & CLIP     & 60.03 & 80.17 & 86.10
\end{tabular}
\caption{The results of the ablation analysis.}
\label{tab:ablation_results}
\end{table}

The first part of the table contains ViT architecture, the second one ConvNeXt~\cite{liu2022convnet}. We have used a base model size for a similar model size. Both models were trained either from scratch or initialized using ImageNet or CLIP~\cite{radford2021learning} pretraining. To see the suitability of pretrained features, models were evaluated directly, and then we trained them using a standard protocol using either triplet, InfoNCE, or ICon loss.
We can see that the results improve when training on sketch-image pairs and when using pretraining, which indicates the advantage of using a larger amount of data and a cross-modal training scenario. As expected, a model trained using triplet loss already achieves similar performance to the early works on scene-level SBIR~\cite{chowdhury2022fscoco}. Using a more efficient loss formulation significantly improves performance. In addition, ConvNeXt models tend to work better than ViT for this task by a large margin.

\subsection{Embedding Space Analysis}
\label{sec:embedding}

\begin{figure}
    \centering
    \includegraphics[width=\linewidth]{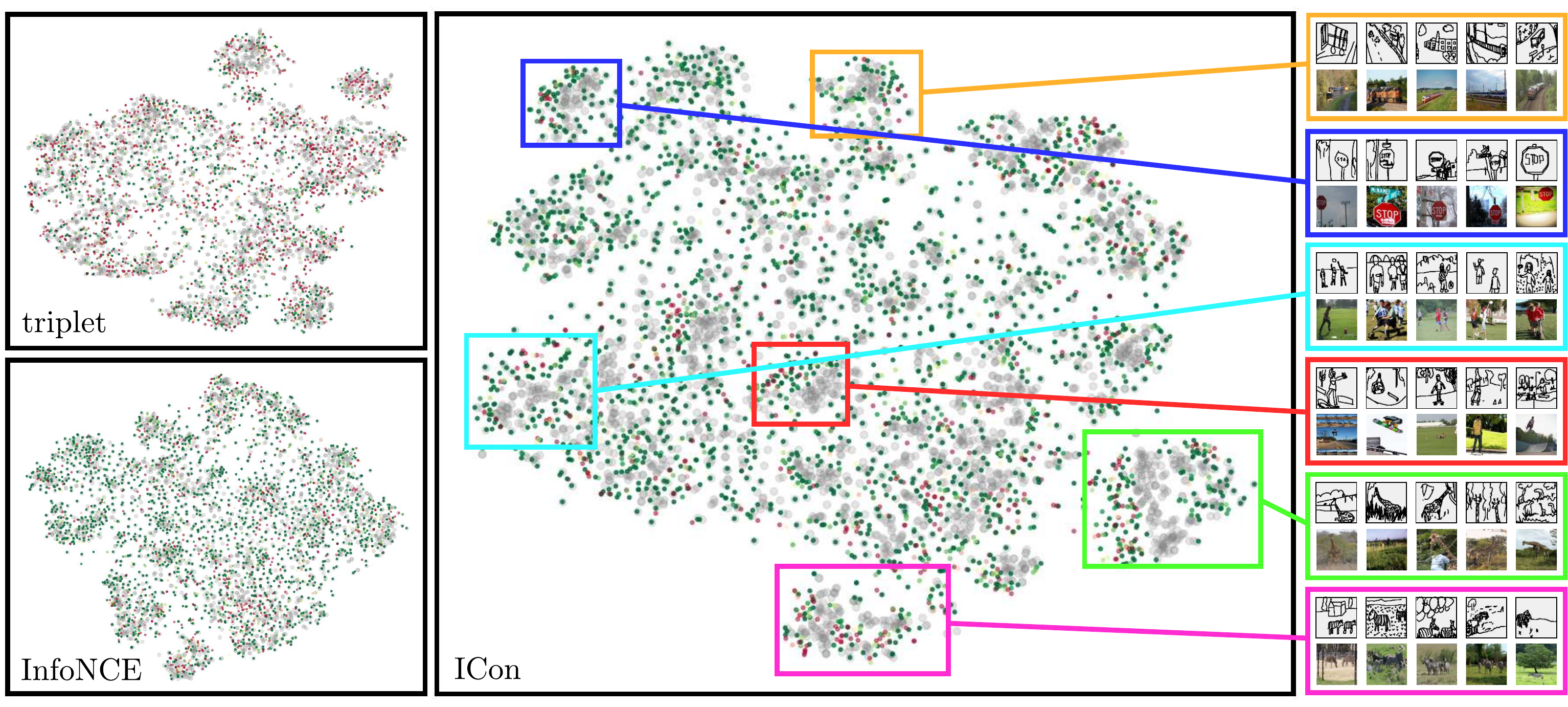}
    \caption{A t-SNE~\cite{van2008visualizing} visualization of embedding space. We use FS-COCO \textit{unseen} test split and compare the space, structured by triplet, InfoNCE, and ICon losses.}
    \label{fig:space}
\end{figure}

In order to better understand the ways different training losses structure the embedding space, we have generated t-SNE~\cite{van2008visualizing} visualizations, seen in Figure~\ref{fig:space}. In the visualization, images are represented with gray dots. Sketches are represented with green and red dots, where green indicates a good match (in the original space) that transitions to red for sketches where the correct photo was not among the closest ten neighbours. We can observe that while some clustering already occurs with triplet loss, it is even more pronounced in the case of InfoNCE and ICon.

For the ICon loss, we also visualize a random selection of sketches and photos of several clusters to validate their content. We can observe that individual clusters indeed contain semantically very similar content (images and sketches) and that certain structure exists beyond individual clusters (human activity clusters are closer together than animal clusters). The difference between the InfoNCE and ICon clusters is that sketches and images within individual clusters overlap much less for ICon. We interpret this as a good thing; the sketches are ambiguous abstractions, therefore, their latent representation should not completely match the corresponding image. We assert that our relaxed loss formulation ensures that the sketches are not aligned completely but still remain sufficiently close to the relevant images.

\subsection{Scalability Analysis}
\label{sec:scalability}

Finally, an often overlooked aspect of SBIR is its scalability. In practical retrieval scenarios, the model would have to deal with a much larger number of images in the database. To simulate this scenario, we augment the test-time gallery with an additional 40,000 images from MS-COCO~\cite{lin2014microsoft} that are not part of FS-COCO~\cite{chowdhury2022fscoco}. These images are incrementally added in steps of 1,000.

\begin{figure}[h!]
    \centering
    \includegraphics[width=\linewidth]{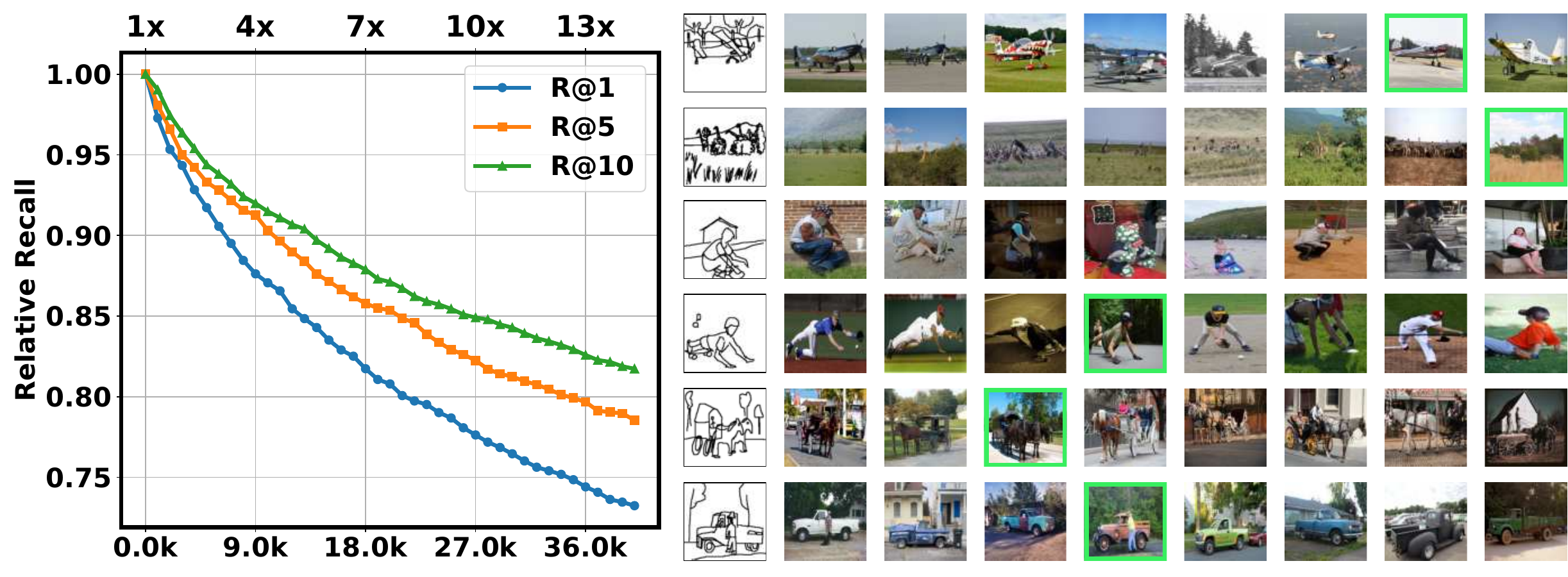}
    \caption{Evaluation of scalability robustness for the proposed model. Left: Relative change to recall values. Right: Examples of images that were ranked above their correct counterparts (marked with a green box).}
    \label{fig:scalability}
\end{figure}

As seen in Figure~\ref{fig:scalability} (left), the R@1 is more sensitive to gallery expansion than R@10, as the inclusion of a single additional image can alter the top-ranked result, whereas R@10 is more robust, allowing multiple ranking changes before the retrieval outcome is affected. Figure~\ref{fig:scalability} (right) shows several examples of where an otherwise perfect match was affected by the addition of new samples. We can see that the model still retains the semantic coherence; however, the ambiguity of sketches means that a more similar image exists.

\subsection{Sketch Ambiguity}
\label{sec:human}

\begin{wrapfigure}{r}{0.4\linewidth}
    \centering
    \includegraphics[width=\linewidth]{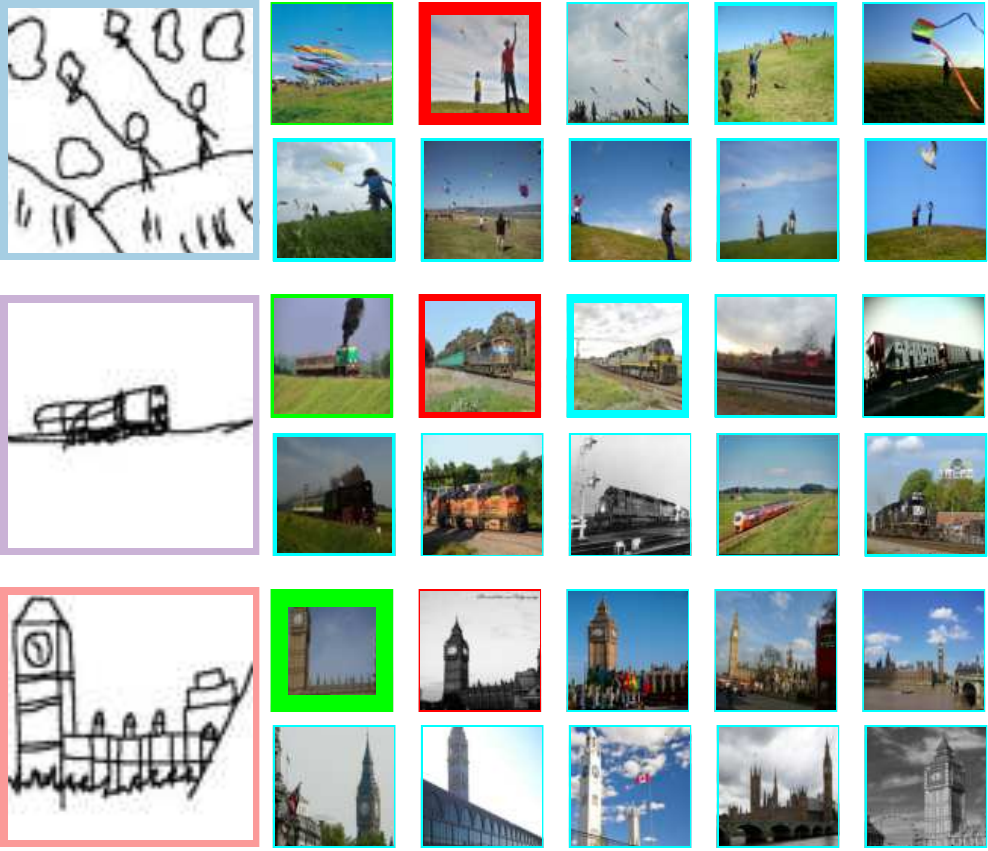}
      \caption{Examples from the sketch ambiguity human survey. The green border indicates a correct match, the red border indicates an incorrect match selected by the model. Border thickness indicates the frequency of selection by participants. \label{fig:ambiguity}}
\end{wrapfigure}

To further confirm the level of ambiguity of sketches, we have selected a subset of sketches and the results proposed by the model, and asked people to select the most appropriate image for a given sketch using an online form. Our analysis revealed three key observations, illustrated by selected examples, shown in Figure~\ref{fig:ambiguity}.

Firstly, in some cases, the majority of participants selected the image ranked highest by our model rather than the ground-truth image (Figure~\ref{fig:ambiguity}, first example). This suggests potential issues with sketch quality, where the model’s top choice may actually be more semantically aligned with the sketch than the ground-truth image, indicating that the sketch may not faithfully represent the intended target. Secondly, several sketches were too vague or abstract for participants to reliably identify the correct image (Figure~\ref{fig:ambiguity}, second example). These cases raise the question of when the model can reasonably be expected to succeed and how to model this correctly. Thirdly, we identified instances where the model correctly ranked the ground-truth image within the top matches, but not as the most similar (Figure~\ref{fig:ambiguity}, third example). In contrast, participants almost always selected the correct image in these cases. This indicates that the model may struggle with capturing fine-grained visual details that humans can easily discern.

\section{Discussion}
\label{sec:discussion}

Our approach offers several practical and methodological advantages for scene-level SBIR. By leveraging pre-trained encoder models and adopting a contrastive objective compatible with large batch training, our method efficiently utilizes available supervision, unlike many prior works that rely on handcrafted sampling strategies or are limited by small batch constraints. Moreover, the design of our training objective explicitly accounts for the inherent ambiguity and noise in sketches, improving robustness to real-world input variability and making the method more generalizable across diverse scene types. 

While we focus specifically on sketch-to-image retrieval, we acknowledge that many recent works explore richer multi-modal narratives combining sketches, text, and different derivatives~\cite{li2023freestyleret}. Although our training formulation is conceptually applicable to other modalities, we leave this extension for future work. Similarly, we do not explicitly address partial or incomplete sketches~\cite{chowdhury2022partially}, which remain an important challenge for practical deployment. On the other hand, we contribute to underexplored areas in the SBIR domain, including the quantification of sketch ambiguity from the human perspective and the scalability of the method in practical scenarios. These aspects are often overlooked in prior art. Given its conceptual simplicity and empirical effectiveness, we believe our method can serve as both a strong baseline and a building block for future developments in scene-level SBIR and broader cross-modal retrieval research.

An important consideration in evaluating SBIR methods is the nature of the available datasets. FS-COCO~\cite{chowdhury2022fscoco} offers diverse and well-annotated samples, but it covers a relatively limited set of scene concepts. While individual categories are well separated, the primary source of retrieval errors arises from ambiguity within semantically similar instances, for example, different variations of a "zebra" or "stop sign" scene. Our relaxed loss formulation helps mitigate some of this ambiguity by softening the supervision signal, but dataset-level improvements are also necessary. Future datasets for sketch understanding should more explicitly account for intra-class ambiguity and scene-level nuance. Incorporating human retrieval performance as part of the annotation process could provide a valuable reference for modeling perceptual similarity and ambiguity in a more grounded way.


\section{Conclusion}
\label{sec:conclusion}

In this paper, we have presented a simple yet powerful method for scene-level SBIR. We have shown that by using appropriate pre-training, architecture, and training loss, a method based on a Siamese encoder may significantly outperform state-of-the-art methods in the field~\cite{li2023freestyleret,liu2022sketcherv2} on a challenging FS-COCO~\cite{chowdhury2022fscoco} dataset as well as on a widely used SketchCOCO dataset~\cite{gao2020sketchycoco}. Beyond the empirical results, our ablation studies reveal several insights into factors influencing performance in scene-level SBIR. We have emphasized the need for well-constructed datasets for the task and demonstrated the problems of the currently available ones. We believe these findings will benefit future work in cross-modal retrieval and contribute to more robust retrieval systems grounded in scene-level understanding.


\vspace{1em}

\noindent \textbf{Acknowledgments}: This research was in part supported by ARIS (Slovenian Research Agency) research programme Computer Vision (P2-0214).

\bibliography{main}

\end{document}